\documentclass{article}
\usepackage{spconf}

\usepackage{color, colortbl}

\usepackage{graphicx}
\usepackage{amsmath,amsfonts,amssymb}
\usepackage{bm}

\usepackage[export]{adjustbox}

\usepackage{svg}

\usepackage{multirow}
\usepackage{makecell}
\usepackage{booktabs}

\usepackage{float}

\usepackage{cite}
\usepackage{url}
\usepackage[hidelinks]{hyperref}
\hypersetup{
    colorlinks=true,
    linkcolor=black,
    filecolor=black,
    urlcolor=gray,
    citecolor=black,
    breaklinks=true
}

\usepackage{tikz}
\usepackage{pgfplots}
\pgfplotsset{width=7cm,compat=1.16}
\usepgfplotslibrary{statistics}
\usetikzlibrary{pgfplots.groupplots}

\usetikzlibrary{plotmarks}
\usetikzlibrary{shapes}

\def\x{{\bm{x}}}
\def\d{{\bm{\delta}}}

\def\L{{\cal L}}

\newcommand{\etal}{\emph{et~al.}}

\definecolor{c1}{RGB}{ 246.2025  131.8860    8.7975}    
\definecolor{c2}{RGB}{   128.0100  128.0100  128.0100}  
\definecolor{c3}{RGB}{          0  202.2405  255.0000}  
\definecolor{c4}{RGB}{          0         0  255.0000}  
\definecolor{c5}{RGB}{          0  255.0000         0}  
\definecolor{c6}{RGB}{          0   87.9240         0}  
\definecolor{c7}{RGB}{          0   0         0}  

\definecolor{Gray}{gray}{0.9}

\usepackage[nomargin,inline,draft]{fixme}
\fxsetup{theme=color,mode=multiuser}
\FXRegisterAuthor{pf}{apf}{\color{blue}pascal}

\title{Improving filling level classification with adversarial training}

\name{Apostolos Modas$^{1}$, Alessio Xompero$^{2}$, Ricardo Sanchez-Matilla$^{2}$, Pascal Frossard$^{1}$, Andrea Cavallaro$^{2}$\thanks{This work is supported by the CHIST-ERA program through the project CORSMAL, under UK EPSRC grant EP/S031715/1 and Swiss NSF grant 20CH21{\_}180444.}}
\address{$^{1}$LTS4, Ecole Polytechnique F\'ed\'erale de Lausanne (EPFL), Switzerland,\\$^{2}$Centre for Intelligent Sensing, Queen Mary University of London, UK}

\begin{document}
\ninept
\maketitle
\begin{abstract}
We investigate the problem of classifying -- from a single image -- the level of content in a cup or a drinking glass. This problem is  made challenging by several ambiguities caused by transparencies, shape variations and partial occlusions, and by the availability of only small training datasets. In this paper, we tackle this problem with an appropriate strategy for transfer learning. Specifically, we use adversarial training in a generic source dataset and then refine the training with a  task-specific dataset. We also discuss and experimentally evaluate several training strategies and their combination on a range of container types of the CORSMAL Containers Manipulation dataset. We show that transfer learning with adversarial training in the source domain consistently improves the classification accuracy on the test set and limits the overfitting of the classifier to specific features of the training data.
\end{abstract}
\section{Introduction}
\label{sec:intro}

The estimation of the amount of content (filling level) within a container is made challenging due to differences in the shape of containers, occlusions caused by the hand holding the container, and transparencies of both the container and the filling (e.g.,~depth estimation may be highly inaccurate for transparent objects~\cite{Sajjan2020ICRA_ClearGrasp}). The few approaches designed to tackle this problem use RGB~\cite{Mottaghi2017ICCV}, thermal~\cite{Schenck2017ICRA}, or a combination of RGB and depth data~\cite{Do2016,Do2018}, and usually observe the action of pouring content in a container over multiple frames~\cite{Schenck2017ICRA,Schenck2017RSS,Do2016,Do2018}. Mottaghi \etal~\cite{Mottaghi2017ICCV} showed that a Convolutional Neural Network (CNN) classifier outperforms a regression model in estimating the filling level using only one RGB image. The best performance was achieved with transfer learning~\cite{TanTransferLearning}: self-collected data were used as task-specific dataset, the {\em target domain}, to fine-tune the parameters of selected layers of the CNN that was previously trained on the much larger ImageNet dataset~\cite{Deng2009CVPR_ImageNet}, the {\em  source domain}.
Transfer learning is suitable for image recognition tasks with only small datasets available for training. Examples of these tasks include fine-grained object classification and scene classification~\cite{huh2016makes,Kornblith_2019_CVPR}, and
the recognition of object properties such as volume, texture, shape and material~\cite{Mottaghi2017ICCV,XueTexture,Farhadi2009,Sajjan2020ICRA_ClearGrasp,Vedaldi2014CVPR}.

\begin{figure}[t!]
    \centering
    \includegraphics[width=.85\columnwidth]{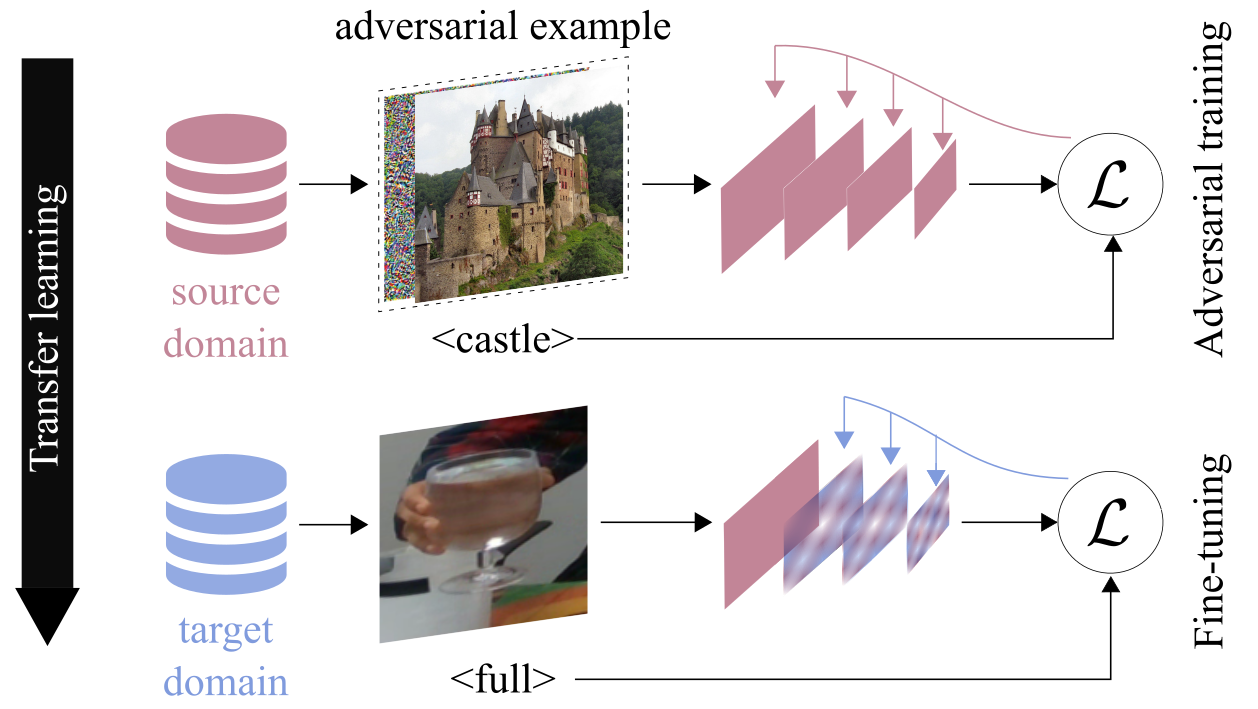}
    \caption{An illustrative 4-layer CNN trained via transfer learning using {adversarial training} on the source domain (adversarial perturbations are added to the original images), followed by fine-tuning some of the layers with images from the target domain. This training strategy achieves the best accuracy in the test set of the experiments. The color of each layer corresponds to the domains used to optimize the classifier parameters using the loss $\mathcal{L}$.
    }
    \label{fig:abstract}
    \vspace{-10pt}
\end{figure}

In this paper, we go beyond  current approaches that use transfer learning to classify the filling level of a container~\cite{Mottaghi2017ICCV}, and evaluate transfer learning combined with adversarial training in the source domain~\cite{salman2020adversarially,utrera2020adversariallytrained} on the small-scale CORSMAL Containers Manipulation (CCM)~\cite{Xompero_CCM} dataset  (Fig.~\ref{fig:abstract}). We thoroughly analyze the performance of standard training, adversarial training, transfer learning, and their combinations, under different setups, which include the number of fixed layers during fine-tuning, and the norm of the perturbation in adversarial training both on the source and the target domain. We show that the generalization of a ResNet-18~\cite{He2016CVPR_ResNet} classifier on the test set of CCM can be limited by its bias towards specific features of the CCM training data. However, adversarially training the classifier on ImageNet, followed by fine-tuning on the train set of CCM, mitigates these biases and consistently produces classifiers with better generalization performance.

\section{Target task and training strategies}
\label{subsec:training_strategies}

  
\definecolor{ts1}{RGB}{ 0  0    0}    
\definecolor{ts2}{RGB}{127 127 127}  
\definecolor{ts3}{RGB}{229, 194, 36}  
\definecolor{ts4}{RGB}{238,113,27}  
\definecolor{ts5}{RGB}{19,219,30}  
\definecolor{ts6}{RGB}{0,58,236}  

We approach the problem of estimating the filling level, $y$, of a container captured in an image $\x\in[0,1]^{H\times W\times C}$, as a classification task, where $H,W,C$ are the height, width, and number of channels respectively. We express the filling level as a percentage of the container's capacity:~$y \in \{0\%,50\%,90\%,\text{\emph{unknown}}\}$, where the \emph{unknown} class helps handling cases with opaque or translucent containers for which the filling level cannot be estimated through direct vision. Let $f_{\theta}$ be a CNN classifier, parameterized by a set of parameters $\theta$, that maps an image $\x$ -- drawn from a distribution $\mathcal{D}$ -- to a label $y$, such that $f_{\theta}(\x)=y$. Given a train set of image-label pairs $\mathcal{T}=\{(\x^i,y^i)\}_{i=1}^N$, the goal is to find a set of parameters that minimizes a suitable loss function $\mathcal{L}(\x,y | \theta)$ such that $f_{\theta}$ correctly predicts $y$ for $\x\sim \mathcal{D}$ but $\x\notin \mathcal{T}$ (generalization).

We refer to the  common strategy for training a classifier on a train set, $\mathcal{T}$, as Standard Training (ST). A good generalization may be achieved if the number of image-label pairs in $\mathcal{T}$ is very large, e.g.,~$N\approx1.2$ millions in ImageNet. However, for the target task of classifying the filling level  such amount of data is not available. Transfer learning helps to overcome this limitation by using an additional training set $\mathcal{S}$, with $|\mathcal{S}|=M \gg N$, that may not be related to the target task.
Transfer learning pre-trains the parameters of $f_{\theta}$ on $\mathcal{S}$ (source domain) and then refines them on $\mathcal{T}$ (target domain) via fine-tuning (FT). We refer to this strategy as ST$\rightarrow$FT. With ST$\rightarrow$FT, the parameters of some layers in the pre-trained model are fixed and FT only refines those of the remaining layers. We will denote with $L$ the number of layers whose parameters are fixed.

Instead of using the original set of images,
Adversarial Training (AT)~\cite{goodfellowExplainingHarnessingAdversarial2015, moosavi-dezfooliDeepFoolSimpleAccurate2016, madryDeepLearningModels2018} uses images modified with carefully crafted noise, known as adversarial perturbation. This noise is specifically designed to change the decision of a classifier~\cite{szegedyIntriguingPropertiesNeural2014,goodfellowExplainingHarnessingAdversarial2015, moosavi-dezfooliDeepFoolSimpleAccurate2016, madryDeepLearningModels2018,ModasSparseFool}. 
Formally, a perturbation $\d$ is added to an image $\x$ in order to maximize the loss function $\mathcal{L}(\x+\d,y | \theta)$ in a given $\ell_p$-ball of radius $\epsilon$ around $\x$~\cite{goodfellowExplainingHarnessingAdversarial2015, madryDeepLearningModels2018}
\begin{equation}
    \begin{split}
        \max_{\d} & \quad \L(\x+\d,y | \theta)\\
        \text{s.t.} & \quad \|\d\|_p\leq\epsilon \\
        & \quad \x+\d \in [0,1]^{H\times W\times C},
    \end{split}
\label{eq:adv_example}
\end{equation}
and the objective of AT is to minimize the adversarial loss $\mathcal{L}(\x+\d,y | \theta)$. The resulting adversarially trained models learn features that correlate better with features of the classes of interest and are thus more robust~\cite{tsiprasRobustnessMayBe2018, allen-zhu_feature_2020,engstrom2019adversarialPriorLearnedRepresentations, ImageSynthesis}.
Hence, $f_{\theta}$ is expected to learn more task-relevant features with  AT. 
While AT was originally designed to increase the robustness of deep networks to adversarial perturbations~\cite{moosavi-dezfooliDeepFoolSimpleAccurate2016, madryDeepLearningModels2018}, it has also contributed to other tasks~\cite{ortizjimenez2020optimism}.
Recently, it was shown that AT in the source domain can improve transfer learning~\cite{salman2020adversarially,utrera2020adversariallytrained}: adversarially trained models  from a source domain can help improving the accuracy on the target task after fine-tuning, despite performing worse, in terms of task accuracy, on the source domain.

We aim to evaluate this training strategy on the filling-level classification task and to compare it against five other strategies. As training strategies we consider ST$\rightarrow$FT~\cite{Mottaghi2017ICCV}; ST on the target domain; AT on the target domain; and three  combinations of AT with transfer learning, namely AT on the \emph{source} domain (AT$\rightarrow$FT), AT on the \emph{target} domain (ST$\rightarrow$AFT), and AT on \emph{both} domains (AT$\rightarrow$AFT). 

AT$\rightarrow$FT adversarially pre-trains the parameters of $f_{\theta}$ on the \emph{source} domain $\mathcal{S}$ and then fine-tunes them on the target domain $\mathcal{T}$~\cite{salman2020adversarially,utrera2020adversariallytrained}. Similarly to what was observed in~\cite{salman2020adversarially,utrera2020adversariallytrained}, we expect that the performance of fine-tuning on $\mathcal{T}$ will further improve if we use a model trained on $\mathcal{S}$ with AT instead of a model trained with ST, even if the classification performance of the robust model on $\mathcal{S}$ is worse than the performance of the model trained with ST. The exact reason behind this improvement is still an open question, but it is related to the differences in the learned features between standard and robust models. Also, this improvement depends on the $\epsilon$ used in Eq.~\ref{eq:adv_example} during AT, and the value that leads to better accuracy may differ across tasks and domains.  Smaller values for $\epsilon$  generally lead to better performance~\cite{utrera2020adversariallytrained}, but its value will be selected empirically.

\begin{figure}[t!]
    \centering
    \includegraphics[width=.9\columnwidth]{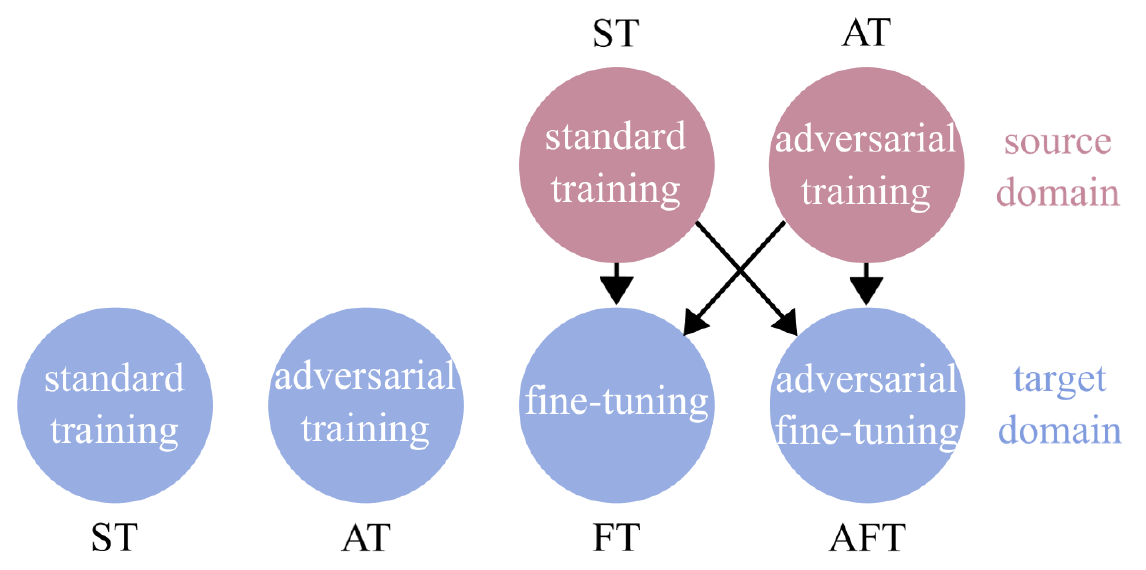}
    \caption{The six training strategies analyzed in our experiments: independent standard training (ST) and adversarial training (AT) on the target domain, and four transfer learning strategies from source to target domain via fine-tuning (FT).
    }
    \label{fig:statediagram}
    \vspace{-5pt}
\end{figure}

Finally, the last two training  combinations apply AT either on the \emph{target} domain (ST$\rightarrow$AFT) via FT or on \emph{both} domains (AT$\rightarrow$AFT). Considering the effect of AT on the features learned by a classifier, we will investigate how $f_{\theta}$ is affected when the transferred learned features from $\mathcal{S}$ are further filtered by AT on $\mathcal{T}$.
Fig.~\ref{fig:statediagram} summarizes the training strategies under analysis, which will be compared in the next section.

\section{Experimental validation}
\label{sec:experiments}

\subsection{Dataset}
\label{sec:dataset}

The task is to classify the filling level from a single RGB image. The  CCM dataset~\cite{Xompero_CCM} comprises of four views capturing under  different backgrounds and illumination conditions cups and drinking glasses. The containers are transparent, translucent or opaque. The content is transparent (water) or opaque (pasta, rice). Each container stands upright on a surface or is being manipulated by a person. We only consider data of the public CCM repository, namely 4 cups and 4 drinking glasses. 
 
\begin{figure}[t!]
    \centering
    \includegraphics[width=.85\columnwidth]{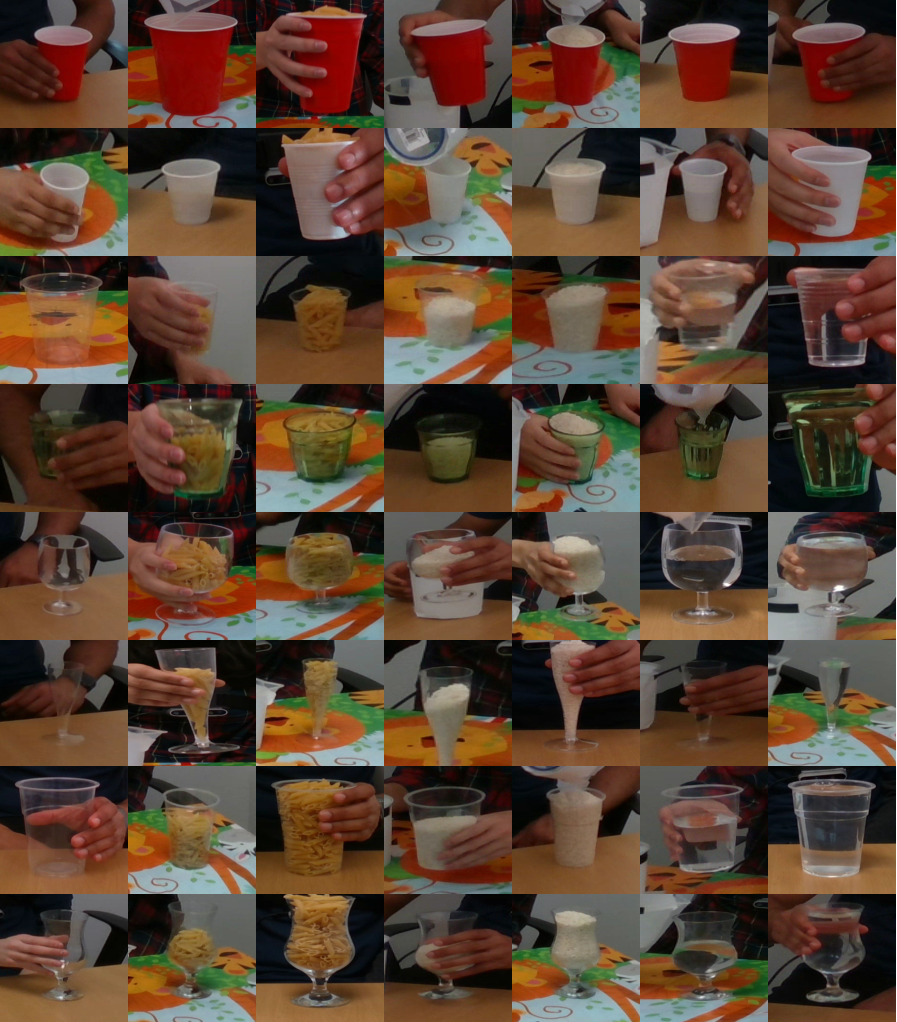}
    \caption{Sample images (resized crops) from the CORSMAL Containers Manipulation dataset~\cite{Xompero_CCM}. Each column shows different filling types and levels, and each row shows different backgrounds and hand occlusions.
    }
    \label{fig:dataset}
    \vspace{-7pt}
\end{figure}

From the CCM video data, we automatically sampled and then visually verified 10,269 frames of containers for which a pouring action was completed.
To increase the variability in the sampled data, we selected frames considering that the container is completely visible or occluded by the person's hand, and under different backgrounds. 
For each frame, the final image is extracted by cropping only the region with the container using Mask R-CNN~\cite{He2017ICCV_MaskRCNN}, followed by visual verification. 
Each crop is associated to an annotation of filling type and filling level (empty or filled at 50\% or 90\% of the  capacity of the container), hand occlusion, and transparency of the container. We call this image dataset Crop-CCM or C-CCM. Sample C-CCM images\footnote{Sampled images can be found at \url{https://corsmal.eecs.qmul.ac.uk/filling.html}} are shown in 
Fig.~\ref{fig:dataset}.

\begin{figure}[t!]
    \centering
    \includegraphics[width=\columnwidth]{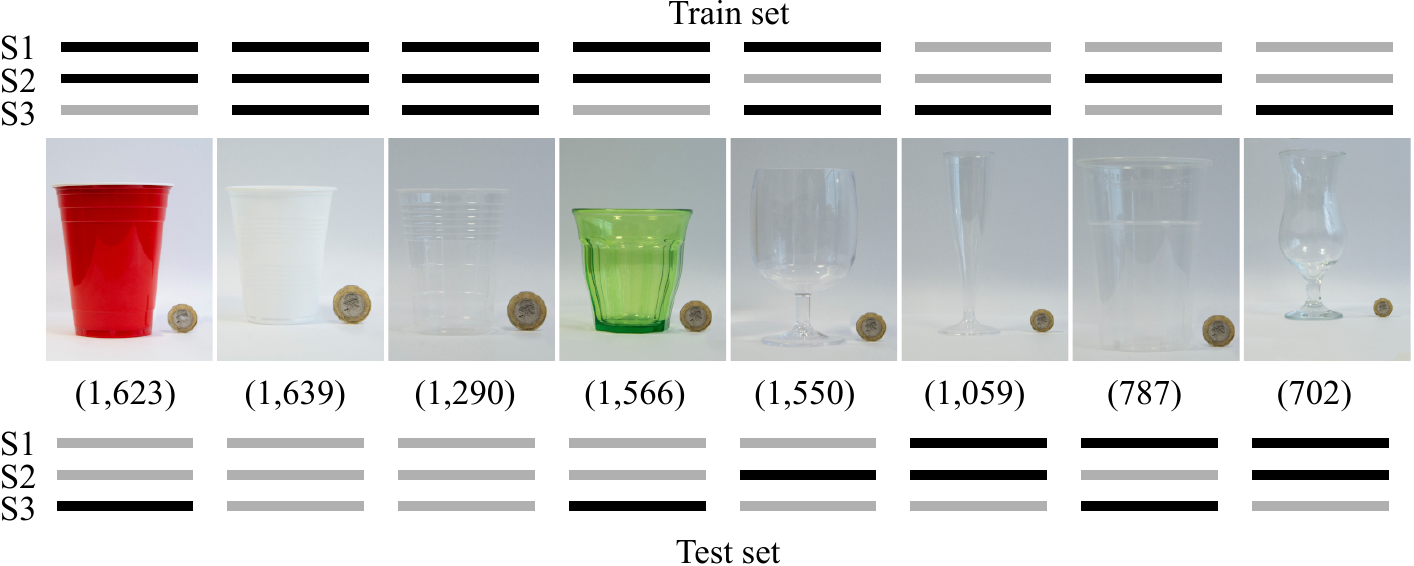}
    \caption{Comparison of three train and test splits (S1, S2, S3) of the public containers from CCM for the shape analysis in the experiments. Black lines mean that the set of images belonging to that container are part of the train (test) set in the data split. The number of images for each container are shown in parentheses. Note the diversity in shape, color, texture, and transparency, as well as the size compared to the 1-pound coin (GBP) used as reference size.
    }
    \label{fig:containers}
    \vspace{-5pt}
\end{figure}

To investigate the impact of the shape of a container on this task, we split C-CCM into train and test sets under three configurations, based on the container type. The first configuration ($\text{S}_1$) considers a {champagne flute} in the test set to further increase the shape variability of containers not previously seen in the train set. The second configuration ($\text{S}_2$) swaps a {beer cup} with a {wine glass} to analyze the influence of the stem of the wine glass. The last configuration ($\text{S}_3$) places all the containers with a stem 
in the train set, and the test set contains only cups without stem. Fig.~\ref{fig:containers} shows the three configurations and the number of samples for each container type. 

\subsection{Classifier and implementation choices}
\label{subsec:implementation}

We use as classifier a ResNet-18~\cite{He2016CVPR_ResNet}. Note that we also conducted experiments using a ResNet-50 and a WideResNet-50~\cite{wide_resnet}, and the findings are similar to the ones of ResNet-18. The discussion of the results will focus on ResNet-18 as it is the least complex network among the three.
With ST we train the classifier on C-CCM, whereas with AT we train the classifier on images modified with $\ell_2$ adversarial perturbations ($p=2$ in Eq.~\eqref{eq:adv_example}) crafted with the $10$-iteration PGD~\cite{madryDeepLearningModels2018}. With the transfer learning  strategies we fine-tune the available pre-trained models on C-CCM: for ST$\rightarrow$FT and ST$\rightarrow$AFT we use the pre-trained model provided by PyTorch~\cite{paszkePyTorchImperativeStyle}, whereas for AT$\rightarrow$FT and AT$\rightarrow$AFT we use the robust models provided by~\cite{salman2020adversarially}.

For each strategy, we train or fine-tune the classifier for $30$ epochs, using a  cross-entropy loss~\cite{Shalev_Shwartz} and stochastic gradient descent. The learning rate for updating the weights is set to $0.1$ when training directly on C-CCM, and $0.005$ when performing transfer learning. The learning rate decays linearly during training. Note that the models we evaluate are the ones obtained at the end of the training epochs (no early-stopping), while for dealing with class imbalances, the training images in a batch are randomly sampled with probabilities that are inversely proportional to the number of images of each class. 


\pgfplotstableread{layers_scenarios.txt}\sensitivitylayers
\pgfplotstableread{epsilon_scenarios.txt}\sensitivityepsilon

\begin{figure}[t!]
    \centering
    \setlength\tabcolsep{2pt}
    \begin{tabular}{cc}
    \begin{tikzpicture}
    \begin{axis}[
        width=.53\columnwidth,
        ymin=45,ymax=85,
        ylabel={Accuracy (\%)},
        tick label style={font=\footnotesize},
        xmin=0.5, xmax=5.5,
        xtick={1, 2, 3, 4, 5},
        xticklabels={0, 1, 2, 3, 4},
        xlabel={$L$},
        label style={font=\footnotesize},
    ]
    \addplot+[solid, color=black, mark=square*, mark options={mark size=2pt,fill=black}] table[x=Layers, y=Scenario1]{\sensitivitylayers};
    \addplot+[solid, color=black, mark=*, mark options={mark size=2pt,fill=black}] table[x=Layers, y=Scenario2]{\sensitivitylayers};
    \addplot[solid, color=black, mark=triangle*, mark options={mark size=2pt,fill=black}] table[x=Layers, y=Scenario3]{\sensitivitylayers};
    \addplot+[only marks, color=red, mark=square*, mark options={mark size=2pt,fill=red}] coordinates {(2, 78.34)
	};
	\addplot+[only marks, color=red, mark=*, mark options={mark size=2pt,fill=red}] coordinates {(2,65.63)
	};
	\addplot[only marks, color=red, mark=triangle*, mark options={mark size=2pt,fill=red}] coordinates {(2,82.32)
	};
    \end{axis}
    \end{tikzpicture} &
    \begin{tikzpicture}
    \begin{semilogxaxis}[
        width=.532\columnwidth,
        ymin=60,ymax=92,
        tick label style={font=\footnotesize},
        xmin=0.005, xmax=1.5,
        xtick={0.01,0.05,0.1,0.5,1},
        xticklabels={.01,.05,.1,.5,1},
        xlabel={$\epsilon^s$},
        label style={font=\footnotesize},
    ]
    \addplot+[solid, color=black, mark=square*, mark options={mark size=2pt,fill=black}] table[x=Epsilon, y=Scenario1]{\sensitivityepsilon};
    \addplot+[solid, color=black, mark=*, mark options={mark size=2pt,fill=black}] table[x=Epsilon, y=Scenario2]{\sensitivityepsilon};
    \addplot[solid, color=black, mark=triangle*, mark options={mark size=2pt,fill=black}] table[x=Epsilon, y=Scenario3]{\sensitivityepsilon};
	\addplot+[only marks, color=red, mark=square*, mark options={mark size=2pt,fill=red}] coordinates {(0.05, 80.97)
	};
	\addplot+[only marks, color=red, mark=*, mark options={mark size=2pt,fill=red}] coordinates {(1,73.27)
	};
	\addplot[only marks, color=red, mark=triangle*, mark options={mark size=2pt,fill=red}] coordinates {(0.5,88.23)
	};
    \end{semilogxaxis}
    \end{tikzpicture}
    \end{tabular}
    \caption{
    Sensitivity analysis for the number of fixed layers $L$ with ST$\rightarrow$FT (left) and for the maximum amount of perturbation bound, $\epsilon^s$, with AT$\rightarrow$FT on test set of the three  dataset splits: first split $\text{S}_1$ (\protect\raisebox{1pt}{\protect\tikz \protect\draw[black,fill=black] (0,0) rectangle (1.ex,1.ex);}), second split $\text{S}_2$ (\protect\raisebox{1pt}{\protect\tikz \protect\draw[black,fill=black] (1,1) circle (0.5ex);}), third split $\text{S}_3$ (\protect\raisebox{1pt}{\protect\tikz \protect\node[fill=black,regular polygon,regular polygon sides=3,inner sep=1.5pt] at (-0.5,0) {};}). Red indicates the highest achieved accuracy.
    Note the different scale of the y-axis, and the logarithmic scale for the x-axis (right).
    }
    \label{fig:analysislayers}
    \vspace{-7pt}
\end{figure}

\pgfplotstableread{accuracy_new.txt}\accscenarios
\begin{figure*}[t!]
    \vspace{0.8cm}
    \centering
    \begin{tikzpicture}
    \begin{axis}[
    axis x line*=bottom,
    axis y line*=left,
    enlarge x limits=false,
    ybar,
    width=\linewidth,
	bar width=3pt,
    xmin=0,xmax=12,
    xtick=data,
    height=0.4\columnwidth,
    ymin=0,  ymax=100,
    ytick={0,20,40,60,80,100},
    ylabel={Accuracy (\%)},
    label style={font=\footnotesize},
    tick label style={font=\footnotesize},
    ymajorgrids=true,
    ]
    \addplot+[ybar, black, fill=ts1, draw opacity=0.5] table[x=CoID,y=S1]{\accscenarios};
    \addplot+[ybar, black, fill=ts2, draw opacity=0.5] table[x=CoID,y=S2]{\accscenarios};
    \addplot+[ybar, black, fill=ts3, draw opacity=0.5] table[x=CoID,y=S3]{\accscenarios};
    \addplot+[ybar, black, fill=ts4, draw opacity=0.5] table[x=CoID,y=S4]{\accscenarios};
    \addplot+[ybar, black, fill=ts5, draw opacity=0.5] table[x=CoID,y=S5]{\accscenarios};
    \addplot+[ybar, black, fill=ts6, draw opacity=0.5] table[x=CoID,y=S6]{\accscenarios};
    \end{axis}
    \begin{axis}[
        axis x line*=top,
        axis y line*=right,
        width=\linewidth,
        height=.4\columnwidth,
        xmin=0,xmax=12,
        tick label style={font=\footnotesize, align=center,text width=3cm},
        xtick={2,6,10},
        xticklabels={$\text{S}_1$,$\text{S}_2$,$\text{S}_3$},
        typeset ticklabels with strut,
        label style={font=\footnotesize},
        ymin=0,ymax=100,
        yticklabels={},
    ]
    \end{axis}
    \node[inner sep=0pt] (flute1) at (1.35,-0.9)
    {\includegraphics[width=.12\columnwidth]{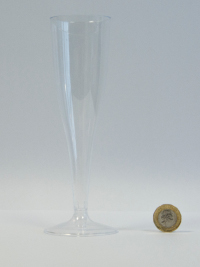}};
    \node[inner sep=0pt] (flute1) at (2.70,-0.9)
    {\includegraphics[width=.12\columnwidth]{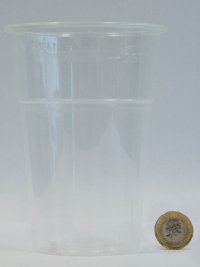}};
    \node[inner sep=0pt] (flute1) at (4.05,-0.9)
    {\includegraphics[width=.12\columnwidth]{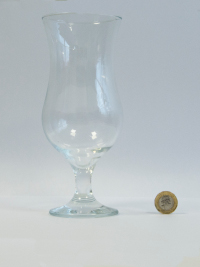}};
    \node[inner sep=0pt] (flute1) at (6.75,-0.9)
    {\includegraphics[width=.12\columnwidth]{champagne_flute_glass.jpg}};
    \node[inner sep=0pt] (flute1) at (8.10,-0.9)
    {\includegraphics[width=.12\columnwidth]{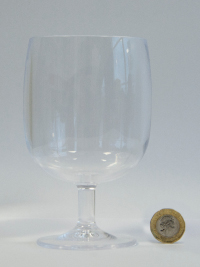}};
    \node[inner sep=0pt] (flute1) at (9.45,-0.9)
    {\includegraphics[width=.12\columnwidth]{cocktail_glass.jpg}};
    \node[inner sep=0pt] (flute1) at (12.15,-0.9)
    {\includegraphics[width=.12\columnwidth]{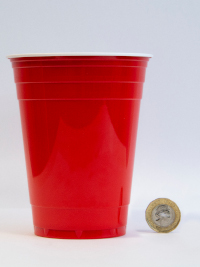}};
    \node[inner sep=0pt] (flute1) at (13.50,-0.9)
    {\includegraphics[width=.12\columnwidth]{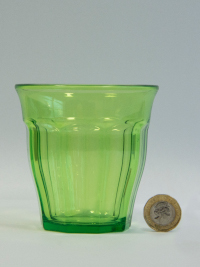}};
    \node[inner sep=0pt] (flute1) at (14.85,-0.9)
    {\includegraphics[width=.12\columnwidth]{beer_cup.jpg}};
    \end{tikzpicture}
   \caption{
   Comparison of the per-container filling level classification accuracy (\%) for the six training strategies. Note the different containers in the test set for each dataset split (see Fig.~\ref{fig:containers} for the train set of each split).
   Legend:
    \protect\raisebox{2pt}{\protect\tikz \protect\draw[ts1,line width=2] (0,0) -- (0.3,0);}~ST,
    \protect\raisebox{2pt}{\protect\tikz \protect\draw[ts2,line width=2] (0,0) -- (0.3,0);}~AT,
    \protect\raisebox{2pt}{\protect\tikz \protect\draw[ts3,line width=2] (0,0) -- (0.3,0);}~ST$\rightarrow$FT,
    \protect\raisebox{2pt}{\protect\tikz \protect\draw[ts4,line width=2] (0,0) -- (0.3,0);}~ST$\rightarrow$AFT,
    \protect\raisebox{2pt}{\protect\tikz \protect\draw[ts5,line width=2] (0,0) -- (0.3,0);}~AT$\rightarrow$FT,
    \protect\raisebox{2pt}{\protect\tikz \protect\draw[ts6,line width=2] (0,0) -- (0.3,0);}~AT$\rightarrow$AFT.
   }
    \label{fig:shapeanalysis}
\end{figure*}

\subsection{Sensitivity analysis}
\label{subsec:sensanalyis}

We perform a sensitivity analysis on the number of fixed layers ($L$)  in  fine-tuning with ST$\rightarrow$FT,  ST$\rightarrow$AFT, AT$\rightarrow$FT and AT$\rightarrow$AFT; and to select the size of the bound for crafting the adversarial perturbation for  AT, ST$\rightarrow$AFT, AT$\rightarrow$FT and AT$\rightarrow$AFT.
Note that we differentiate the bound $\epsilon$ for the  source, $\epsilon^s$, and target, $\epsilon^t$, domain. 
Specifically, we perform the sensitivity analysis only for $\epsilon^s$ with AT$\rightarrow$FT, and for each data split configuration we select the $\epsilon^s$ for which AT$\rightarrow$FT achieves the highest accuracy.
Then, based on these values of $\epsilon^s$, for each data configuration we set $\epsilon^t=\epsilon^s$: since we use $10$-iteration $\ell_2$-PGD~\cite{madryDeepLearningModels2018}, performing a sensitivity analysis or a grid search on $\epsilon^t$ is computationally inefficient, as it is analogous to increasing almost $10\times$ the training epochs.

We first analyze the classification accuracy on the test sets of the three dataset splits when varying the number of fixed layers for ST$\rightarrow$FT as $L=\{0,1,2,3,4\}$. Note that for a ResNet-18 classifier, a layer is a ResNet block of convolutions and batch normalization (see the original ResNet paper~\cite{He2016CVPR_ResNet}).
Since the target dataset is small, it is reasonable to fix the first layer ($L=1$) in order to prevent the classifier from a possible overfitting~\cite{Yosinski2014}.
Indeed, Fig.~\ref{fig:analysislayers} (left) shows that the accuracy on the test set of all configurations (S1, S2, S3) is consistently higher for $L=1$ (78.34\%, 65.63\%, 82.32\%), while it gradually decays as $L$ grows. This is also expected~\cite{Yosinski2014}, since we allow fewer layers to be fine-tuned on the target datasets, and the classifiers then mostly use fixed features from ImageNet. Therefore, we set $L=1$ for ST$\rightarrow$FT as well as for ST$\rightarrow$AFT, AT$\rightarrow$FT, and AT$\rightarrow$AFT.

By fixing $L=1$, we analyze the classification accuracy of AT$\rightarrow$FT when varying the size of the adversarial perturbation on the source domain, $\epsilon^s$. Fig.~\ref{fig:analysislayers} (right) shows that the highest achieved accuracy is different for each dataset configuration: 80.97\% for $\text{S}_1$ with $\epsilon^s=0.05$, 73.27\% for $\text{S}_2$ with $\epsilon^s=1$, and 
88.23\% for $\text{S}_3$ with $\epsilon^s=0.5$. As mentioned in Sec.~\ref{subsec:sensanalyis}, we use these values $\epsilon^s$ also for $\epsilon^t$ when performing AT, ST$\rightarrow$AFT, and AT$\rightarrow$AFT. However, we observed that the model trained with ST$\rightarrow$AFT is unable to converge (train accuracy around 45\%) on $\text{S}_2$ for $\epsilon^t=1$ and on $\text{S}_3$ for $\epsilon^t=0.5$, while it successfully converges on $\text{S}_1$ for the smaller $\epsilon^t=0.05$. We believe that this might be caused by the fact that AT with larger $\epsilon^t$ values eliminates many non-robust, yet useful, features transferred from ImageNet, and prevents the model from fitting the remaining features. Hence, we  set $\epsilon^t=0.05$ for ST$\rightarrow$AFT across all dataset configurations for the rest of the experiments.

\subsection{Results}
\label{subsec:results}

Fig.~\ref{fig:shapeanalysis} shows the filling level classification performance on the three configurations,  $\text{S}_1$,  $\text{S}_2$ and  $\text{S}_3$, for all the training strategies. Constrained by the amount, and hence by the diversity, of training images, the differently trained classifiers could potentially develop biases or overfit to some features, such as the shape of a container. AT$\rightarrow$FT achieved superior performance most of the times. With transfer learning, the features introduced from ImageNet (source domain) appear to decrease such biases, and enable the classifiers to identify features in the train set that are more generalizable. When combining transfer learning with AT at the source domain, the biases are modulated with the transferred features that are also filtered by AT, and the generalization of the classifier further increases. These results confirm that adversarial training improves transfer learning, even in the context of the challenging filling level classification task.

Overall, whenever the performance of ST is low, all transfer learning strategies lead to a significant improvement. On the contrary, whenever ST performs well, the contribution of transfer learning is insignificant, and sometimes it even decreases the final performance. Furthermore, applying AT on the \emph{target} domain, either alone or combined with transfer learning, may even be harmful for the classifier.  

For $\text{S}_1$, the accuracy of ST on the beer cup (middle) is already very high, and the other training strategies do not further improve it. This might be explained by the similar shape of the small transparent cup in the training set. On the other hand, the accuracy on the cocktail glass (right) is similar for all strategies, but lower than the one of the beer cup, with AT$\rightarrow$FT performing slightly better than the rest of the training strategies. Although there is another container with a stem in the training set (wine glass), these accuracy levels might be due to the different shape above the stem that the cocktail glass has, compared to the wine glass. As for the champagne flute (left), the performance of ST and AT is quite low ($\sim$46\%), which might be caused by the unique shape of the flute (narrowing towards the bottom) with respect to the shapes in the training set. However, the accuracy significantly improves with transfer learning.  Especially AT$\rightarrow$FT outperforms all the other strategies by $\sim$30 percentage points (pp). 

For $\text{S}_2$, the accuracy of all strategies on the champagne flute (left) is similar to the one achieved on $\text{S}_1$. The accuracy on the cocktail glass (right) is much lower for most strategies ($\sim$10pp less compared to the performance on $\text{S}_1$), except AT$\rightarrow$FT, which drops only by 3pp and again outperforms the rest of the strategies. The drop of the other strategies could be caused by the lack of a container with a stem in the training set. Finally, the performance on the wine glass (middle) is similar for most strategies, with AT$\rightarrow$FT being again slightly better than the rest. Compared to the cocktail glass, the higher accuracy of all strategies on the wine glass could be caused by the similarity of its shape above the stem with the other transparent cups in the training set, despite the fact that no container with a stem is presented in the training set.

For $\text{S}_3$, the accuracy of ST on the beer cup (right) is high and the other training strategies do not improve it. Instead, the accuracy of ST on the green glass (middle), which has a different shape, is lower and reaches an accuracy of 66\%. However, although ST$\rightarrow$FT does not improve the accuracy, AT$\rightarrow$FT significantly increases it (almost 10pp). The red cup (left) obtains the most interesting improvement compared to the 0.005\% accuracy of ST: all transfer learning techniques achieve an accuracy above 90\%, with AT$\rightarrow$FT achieving 99.5\% classification accuracy. By inspecting the predictions of ST and AT, the classifier assigned the label full (filling level: $90\%$) almost 99\% of the times. In fact, predicting the $\text{\emph{unknown}}$ class is conceptually different from estimating the filling level, and it is more related to classifying non-transparent containers. In this sense, the features learned for transparent objects that are full with rice or pasta might be correlated with the features of the red cup.

\section{Conclusion}
\label{sec:conclusion}

We investigated how different training strategies impact the classification of the filling level of a container. 
Using adversarial training on the source dataset, ImageNet, followed by transfer learning on the target dataset, selected from the CORSMAL Containers Manipulations dataset, permits to consistently improve generalization to unseen containers. Our analysis demonstrates the possibilities of exploiting adversarial training for tasks that extend beyond classical image classification settings. As future work, we will explore other sources of biases that might be related to transparencies, occlusions, or the content, we will investigate alternative ways to avoid overfitting to features of the training data, and we will extend our analysis to other datasets and settings.

\bibliographystyle{IEEEbib}
\bibliography{strings,refs}

\end{document}